# Evolution of Efficient Symbolic Communication Codes

Anton Kolonin[1,2][0000-0003-4180-2870]

[1] Novosibirsk State University, Novosibirsk, Pirogova 1, 630090, Russia
[2] Aigents, Novosibirsk, Pravdy 6-12, 630090, Russia
akolonin@gmail.com

**Abstract.** The paper explores how the human natural language structure can be seen as a product of evolution of inter-personal communication code, targeting maximization of such culture-agnostic and cross-lingual metrics such as anti-entropy, compression factor and cross-split F1 score. The exploration is done as part of a larger unsupervised language learning effort, the attempt is made to perform meta-learning in a space of hyper-parameters maximizing F1 score based on the "ground truth" language structure, by means of maximizing the metrics mentioned above. The paper presents preliminary results of cross-lingual word-level segmentation tokenization study for Russian, Chinese and English as well as subword segmentation or morpho-parsing study for English. It is found that language structure form the word-level segmentation or tokenization can be found as driven by all of these metrics, anti-entropy being more relevant to English and Russian while compression factor more specific for Chinese. The study for subword segmentation or morpho-parsing on English lexicon has revealed straight connection between the compression been found to be associated with compression factor, while, surprising, the same connection with anti-entropy has turned to be the inverse.

## 1 Introduction

The latest advances in natural language processing demonstrated by so-called large language models (LLM) [1] are typically based on tokenization relying on hardcoded punctuation rules and subword segmentation based on so called byte-pair encoding (BPE) [2] and different variations of it such as BPE with dropout [3] and dynamic programming encoding (DPE) [4]. Relying on a hardcoded punctuation may be thought as a not fair approach to learn the language bottom-up from the ground. In turn, the known subword segmentation techniques appear not quite conforming to true language morphology. Moreover, the LLMs are based on non-interpretable distributed representations where the interiors of the model can not be explicitly validated to conform with true knowledge are grammatical rules even though the latest versions of can approximate human language decently. The computational resources required for that, given the number of parameters in the latest models is enormous, which may be compensating inaccuracy of the subword tokenization schemes.

The alternative language learning approach based on interpretable formal language models such as Link Grammar [5] has beeb explored in earlier works with some promising results obtained for English language [6], however this work has been based on hardcoded tokenization rules as well.

The attempts on learn tokenization models unsupervisedly have been made in other prior work such as [7] and then [8], where the latest work has been involving multiple languages. Unfortunately, the levels of accuracy presented the latter works have been found way below the ones obtained with hardcoded tokenization schemes. The latest work [9] improves the accuracy of cross-lingual tokenization reaching F1 score 0.99 for English, 1.0 for Russian, and 0.71 for Chinese. Such level of accuracy has been achieved using so-called "transition freedom" ("freedom of transition") metric, apparently relying on the fundamental ground of so-called "free energy principle" suggested by earlier fundamental work [10] where the minimization of uncertainty is posed as a key principle of brain function and hence may be applied to nature of the human language structure. However, the latest results have been achieved with manual search for optimal hyper-parameters. The very latest work [11], applied for the same three languages, has attempted to find a way for meta-learning for optimal hyper-parameters finding connection between target F1 score of tokenization itself and introduced culture-agnostic metrics called normalized anti-entropy (1), compression factor and cross split F1-score described in the paper.

$$\tilde{S} = 1 - H/(log2(L)) \quad (1)$$

The normalized anti-entropy" $\tilde{S}$ defined above based on $H$ is a Shannon entropy of entire training set tokenized with given tokenization model, where $L$ is size of the lexicon underlying the tokenization model.

The compression factor $C\%$ is asserted as the ratio between numerator as "compressed" size of training set given current tokenization model and denominator as uncompressed size of training set. The compressed size is evaluated as length of sequence of token indexes in tokenized text corpus entirely plus size of the "dictionary" - sum of lengths of all tokens in it. The uncompressed size of training set is evaluated just as count of symbols in it.

The cross-split $F1$ score called $CSF1$ in [11] is defined as follows. First, we split the training set corpus in two pieces of the same size, call them set $A$ and set $B$. Next, we create the graph traversal models across N-grams according to the previously cited work [9] for each of the corpus, call them $M(A)$ and $M(B)$. Then, we tokenize the test set with both models, so that $T(M(A))$ and $T(M(B))$ are obtained. Finally, evaluate the cross-split $F1$ score of tokenization as $CSF1$ for $T(M(A))$ against $T(M(A))$ having one as a "ground truth" for another.

Unfortunately, the corpora in the latter work were not well aligned so the conclusions drawn from that work might be not seen as quite reliable, so in this work we have tried to reproduce the claimed results with different aligned corpora for the same English, Russian and Chinese languages. Moreover, we have tried to expand the scope of the study to address the subword segmentation problem.

# 2 Approach

## 2.1 Tokenization or word-level text segmentation

The interpretable language learning developed through earlier works [5,6,9,11] is based on assumption of possibility of learning graph-based models through graph analysis of hierarchy of linguistic entries. The basic parts of the learning process are a) segmentation of stream of smaller linguistic entities into larger groups and b) clustering of these entities on different levels into categories. That is, the former enables learning word-level and punctuation-level tokenization, morphological parsing and then phrase and sentence boundary detection. In turn, the latter makes it possible to learn categories of letters and punctuation such as vowels, consonants, digits, delimiters and quoting symbols or categories of words such as determiners, nouns and verbs.

According to studies in [7,9,11], initial raw representation of the corpus can be built a weighted graph of transitions through the training corpus based on N-grams of different arity. This makes it possible to build further probabilistic models of transitions from any path on the graph to the next segment of the path, having the probabilities and transition freedom computed on each possible transitions. On the level of character-based graph (N=1 for N-gram to N-gram transitions), clustering characters in space of the forward and backward transitions on the graph, only this raw model makes it possible to learn impressive clusters for parts of speech and punctuations as shown on Fig. 1, based on RusAge corpus (`https://www.kaggle.com/datasets/oldaandozerskaya/fiction-corpus-for-agebased-text-classification`).

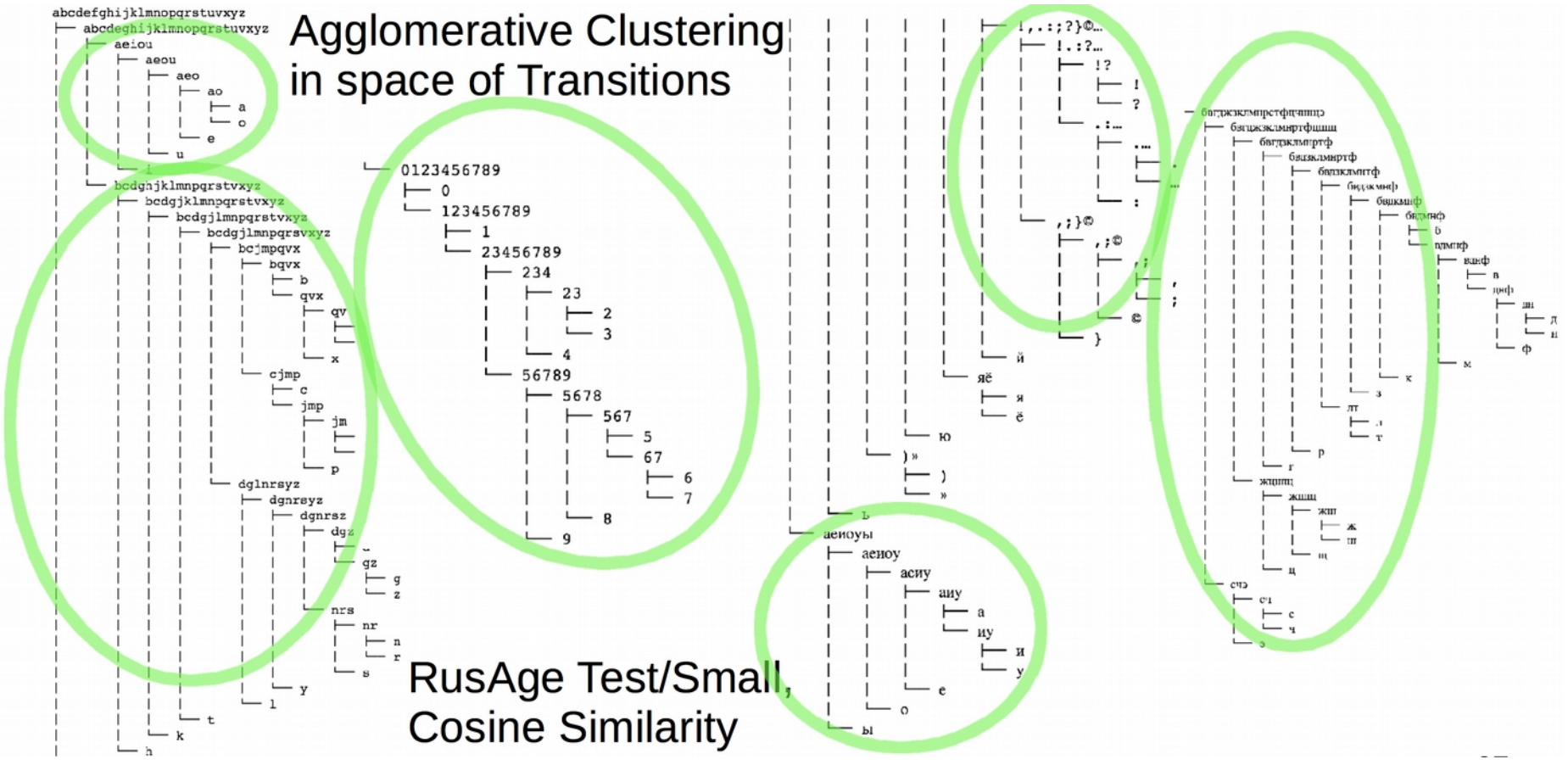

**Fig. 1.** Clustering characters ins space of forward and backward unigram-to-unigram transitions in based on test subset of RusAge corpora. Clusters that may be identified left-to-right: English vowels, English consonants, digits, Russian vowels, punctuation symbols, Russian consonants.

The tokenization technique based on such representations, according to [7,9,11] can be implemented based on detecting the peaks of the transition freedom profiles built on forward and backward traversal along the stream of N-grams. The hyper parameters of such process, according to [9,11] are *N* (oder of N-gram), threshold used to detect the peaks on transition freedom profiles, and threshold to prune low-frequency transitions in the raw model before the tokenization process.

The latest work [11] claims correlation between tokenization *F1* score obtained with different combinations of the hyper-parameters and suggested culture-agnostic metrics as it has been explored my means of grid search such as shown on Fig. 2, applied to English Brown corpus as a train (`http://www.sls.hawaii.edu/bley-vroman/brown_nolines.txt`) and English MagicData corpus as a test (`https://magichub.com/datasets/chinese-english-parallel-corpus-finance`).

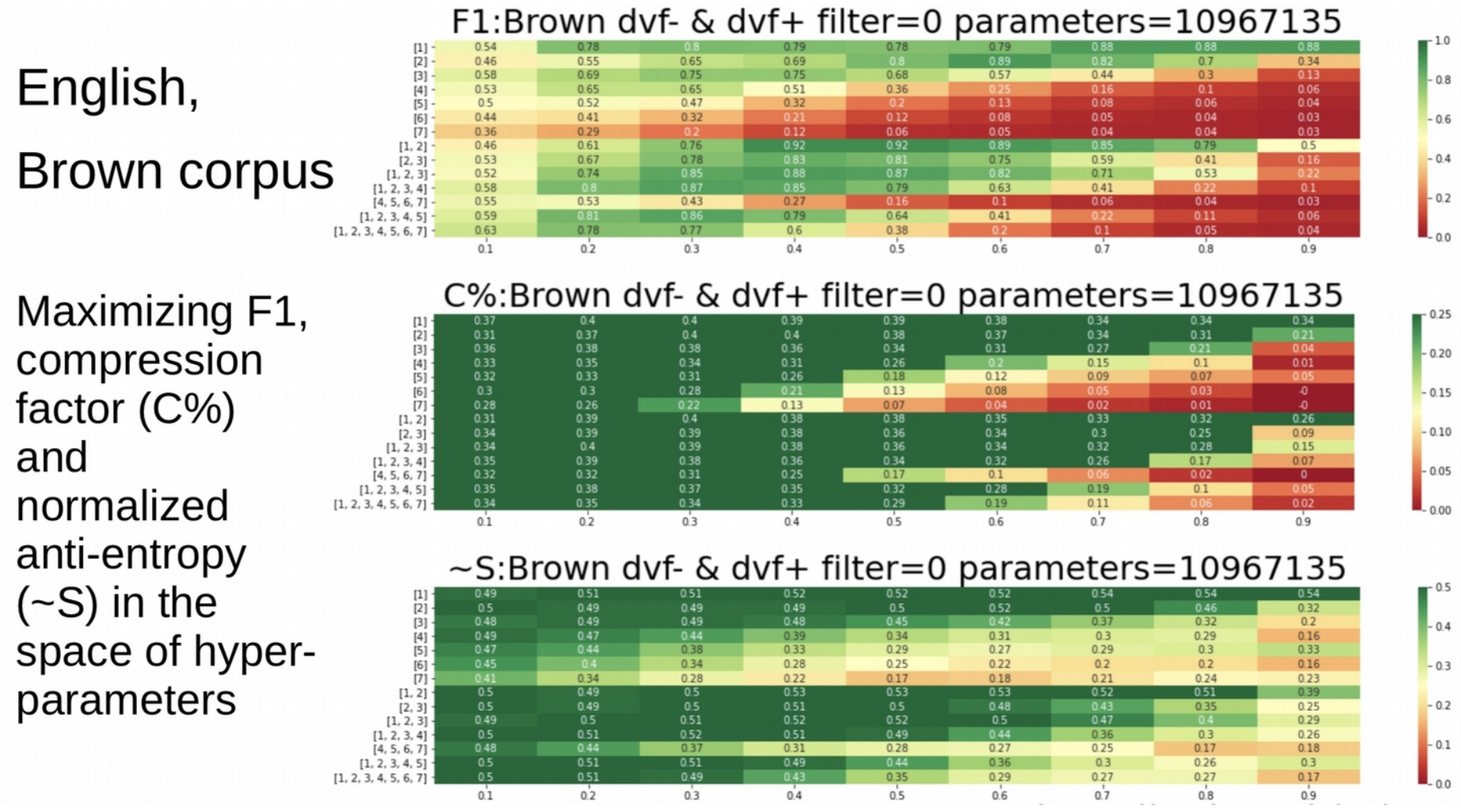

**Fig. 2.** Grid search for the maximized tokenization F1 score (top) along with the best compression factor (middle) and normalized anti-entropy (bottom), correlation is seen across all three.

In this work we attempt to validate the findings mentioned above using technique described in [9,11] but applying to better aligned, richer and diverse train and test corpora.

## 2.2 Morphological parsing or subword-level text segmentation

Moreover, we explore the possibility of subword segmentation or morphological parsing ("morpho-parsing") conducted following the same approach applied to individual pre-segmented tokens or words in attempt to detect word-pieces complying to morphology known to a human language. The primary goal was to see if the described approach can be used to achieve subword segmentation emitting word pieces closer to known morphological units, compared to known approaches [2,3,4]. The secondary goal was to explore connection between the accuracy of such process compared to the culture-agnostic metrics used in the former study.

For initial experiments we have used Aigents English lexicon downloaded from `https://raw.githubusercontent.com/aigents/aigents-java/master/lexicon_english.txt` as a training corpus, where the N-gram-to-character transition graph model has been built for *N* in range from *1* to *7* inclusively, with account to word frequency count, so the transition count per graph edge was taking the word frequency in account.

For reference "ground truth" morphological parsing to compute F1 score against we have used "greedy parser" based on morphological units, such as prefixes and suffixes, downloaded from `https://github.com/aigents/pygents/tree/main/data/corpora/English/morphology`. To control the quality of such reference we have used the control set of words used for illustration in [4] and have got the result of its reference tokenization reviewed by native English speaker with 100% acceptance.

# 3 Results

## 3.1 Tokenization or word-level text segmentation

The first round of unsupervised cross-language tokenization experiments was performed, against two completely different lines of data sets.

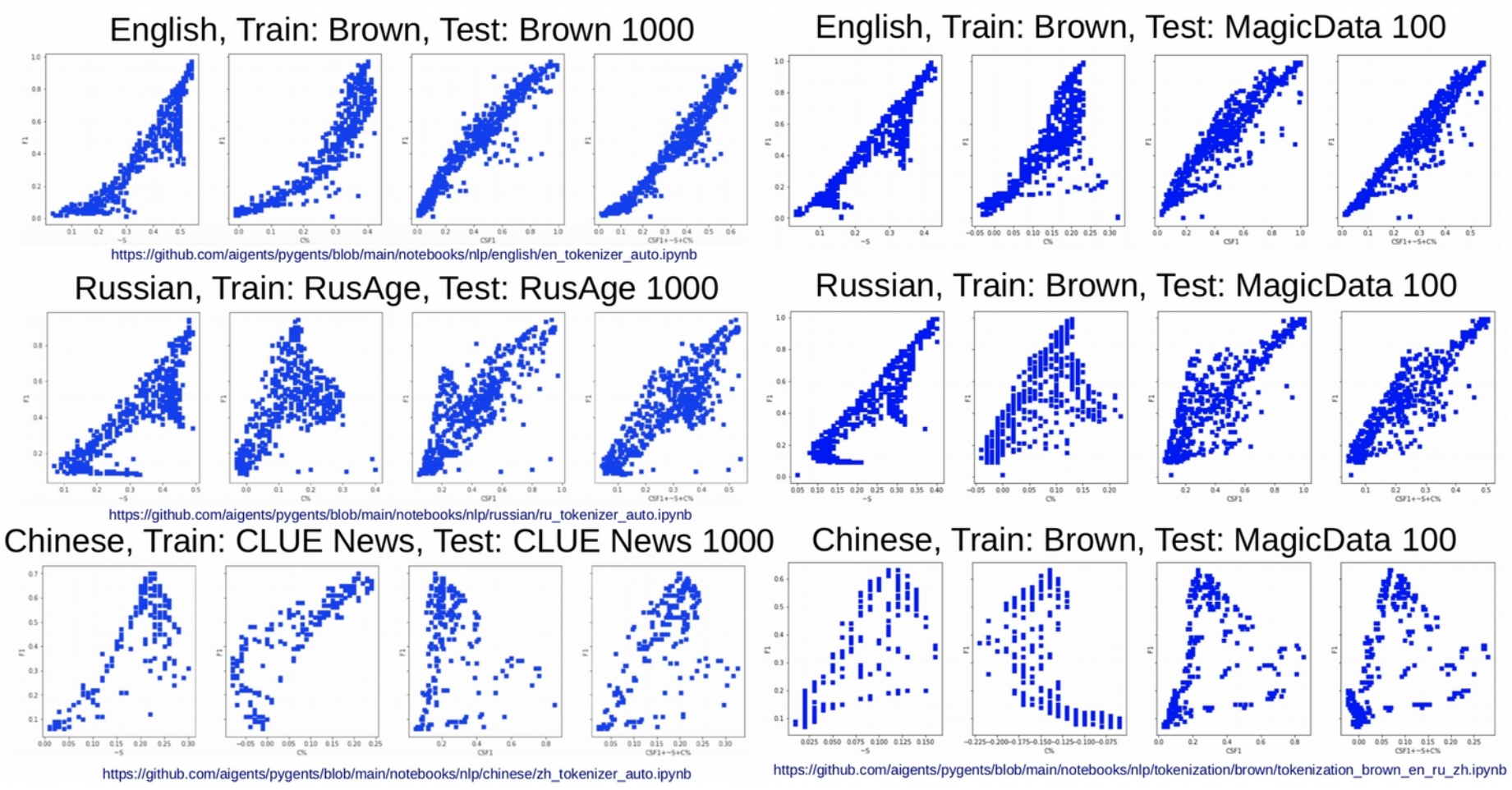

**Fig. 3.** Scatter plots indicating connections of hyper-parameters of unsupervised tokenization with F1 score (vertical axis) and culture-agnostic metrics such as anti-entropy, compression factor, cross-split F1 and average of all three (four plots left to right on each halves of the figure) – for two lines of data sets on the left (non-aligned) and on the right (completely aligned). Results are presented for different languages: English (top), Russian (middle), and Chinese (bottom).

The first line of experiments involved the same corpora as in [9,11] – English Brown, Russian RusAge and Chinese CLUE Benchmark News 2016 dataset (`https://github.com/brightmart/nlp_chinese_corpus`) with 1000 text rows from the every set randomly selected for test, as presented on the left side of Fig.3. Notably, all there corpora were quite different and not aligned in any way, so the obtained results might seem not comparable across the languages. The second line of experiments, presented on the right of Fig.3, involved the same English Brown corpus, machine-translated by Google into Chinese and Russian for training and independent parallel Magic Data corpus for test, so both train and test sets were 100% aligned. The figure shows linear dependency between F1 score and all four culture-agnostic metrics for Russian and English regardless data set however for Chinese this correlation appears less consistent and obvious.

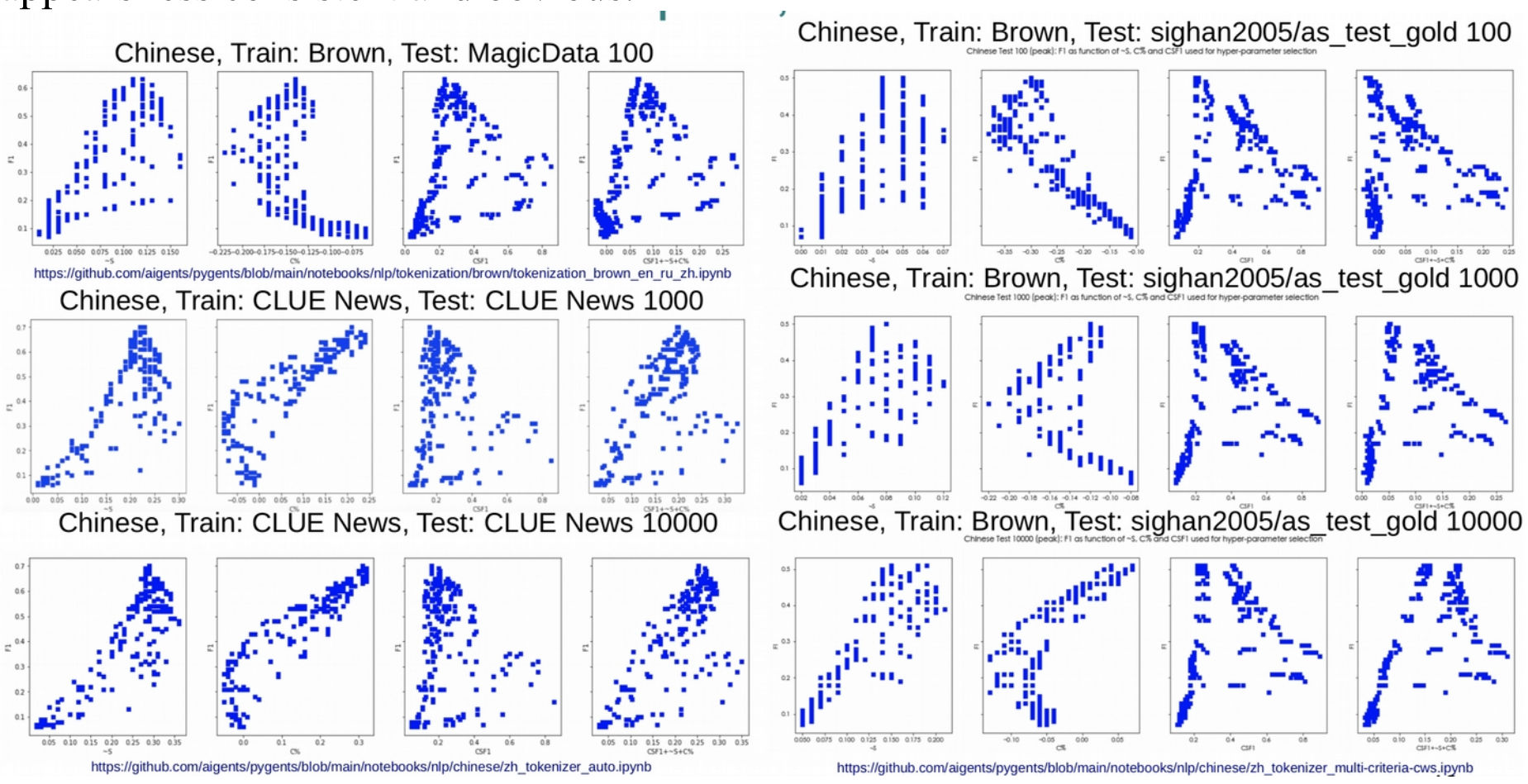

**Fig. 4.** Scatter plots indicating connections of hyper-parameters of unsupervised tokenization with F1 score (vertical axis) and culture-agnostic metrics such as anti-entropy, compression factor, cross-split F1 and average of all three (four plots left to right on each halves of the figure) – for the same Brown data set used for training, but different test sets and testing methodologies and test set sizes. Different test sets: MagicData and CLUE News (left half) and sighan2005/as_test_gold (right half). In the first case (left) reference Jieba tokenizer was used and in the second case (right) manual tokenization markup from the test corpus was used. Different numbers of text lines were used: 100 lines (top), 1000 lines (middle), and 10000 lines (bottom).

Another experiment presented on Fig. 4 has been run on the graph model learned from the same Chinese version of Brown corpus, with different numbers of lines in test sets – 100, 1000, and 1000, sparsely selected from the original test sets. Also, the two different test sets and testing methodologies were used. First, we have used Jieba tokenizer applied to MagicData and CLUE News test sets as in [9,11] and next we have used reference manual tokenization coming as part of the sighan2005/as_test_gold downloaded from `https://github.com/hankcs/multi-criteria-cws/tree/master/data/sighan2005`.

The findings of the latter experiment is that the correlation between F1 score of tokenization and culture-agnostic metrics for Chines becomes obvious and larger volume of testing data, at 10000 lines as shown on Fig. 4. Specifically, anti-entropy and compression factor appear the most clearly correlated with the F1 score. In turn, the cross-split F1 score appear the least clearly correlated being obscured by multiple dots in the right side of each plot, likely corresponding to local extremums in the space of hyper-parameters.

Summarizing the experiments above extending the earlier works [9,11] along with presented on Fig. 2, Fig. 3 and Fig. 4, we can conclude that the structure of the language from tokenization perspective, regardless of specific linguistic culture, at least in context of written English, Russian and Chinese, is optimized to maximize both anti-entropy and compression factor for overall volume of communication.

The above can be explained as follows. For any language, we have the same large corpus of texts, which can be tokenized in different ways, relying on N-gram-to-character transition models built from the corpus, given the hyper-parameters discussed in [9,11]. The set of the hyper-parameters, making it possible to get the most accurate tokenization with superior *F1* score is assumed to correspond to the some cognitive settings in human brain making it possible for humans to comprehend the languages in the way we all do. We can find these hyper-parameters trying tokenizations with different combinations of them in the space of hyper-parameters, referring to test sets with known tokenizations as a "ground truth". But let us explore if the same hyper-parameters can be found without of knowing the "ground truth" as a reference. Can we just pretend the speech and text segmentations are made not just to make the discovered tokens identical to known words given as a reference, but to make the compression of the information stored in the texts efficient and have its entropy minimized. To check this, we can do the tokenization of the same text test sets in the same space of hyper-parameters, computing the compression factor *C%* and anti-entropy $\tilde{S}$ on the tokenized test set [11]. So we try it and we do find that the same hyper-parameters that correspond to the highest compression factor and anti-entropy are also corresponding to the highest *F1* score of tokenization. Moreover, we find the same connection for English, Russian and Chinese.

The other culture-agnostic metric, after compression factor and anti-entropy is cross-split *F1* score described above, referring to [11]. The nature of this metric is the following. Let say we split the same language corpus in two pieces. We can pretend the two pieces are corresponding to different groups of people using the same language to communicate in different patterns and on different aspects, so think of them as a two sub-cultures of the same linguistic culture. Then let us build tokenization model from the first corpus and use the model to tokenize test set from the second corpus. After that, do the opposite – build tokenization model from the second corpus and use it to tokenize test set from the first corpus. For each tokenization of the test sets we compute *F1* score and get average of the two – that is what we call cross-split *F1* score or *CSF1*. We assume that the ability to get higher *CSF1* scores correspond to higher ability of a linguistic sub-culture learned on one subset of language understand texts from another subset of language attributed to another linguistic sub-culture. We can do that with different hyper-parameters as we did it for *C%* and *˜S* against *F1* score.

On the path of this exploration, we have found that the cross-split *F1* score *CSF1* has connection to tokenization *F1* similar to its connection with compression-factor and anti-entropy. So we conclude that the former cross-lingual metric, representing ability of members of different linguistic sub-cultures to understand each other can be considered as another objective for unsupervised identification of the optimal tokenization hyper-parameters for any language. Should be noted that for Chinese this connection appears not as reliable as it is for English and Russian.

The difference between the three languages is that the expression of connection between the metrics apparently depends on the size of alphabet. English with 26 letters has the most expressive connection, see top row of Fig. 3. The expression is more vague for Russian with 33 letters in alphabet, see middle row of Fig. 3. The most blurred connection is for Chinese, see bottom row of Fig. 3. We can attribute it to huge size of Chinese alphabet. Still, the connection between the metrics gets clearer with increase of size of Chinese test set, as seen on Fig. 4, where linear correlation between *C%*, *˜S* and *F1* score across different sets appears with test set size of 10000 lines, see bottom row of Fig. 4.

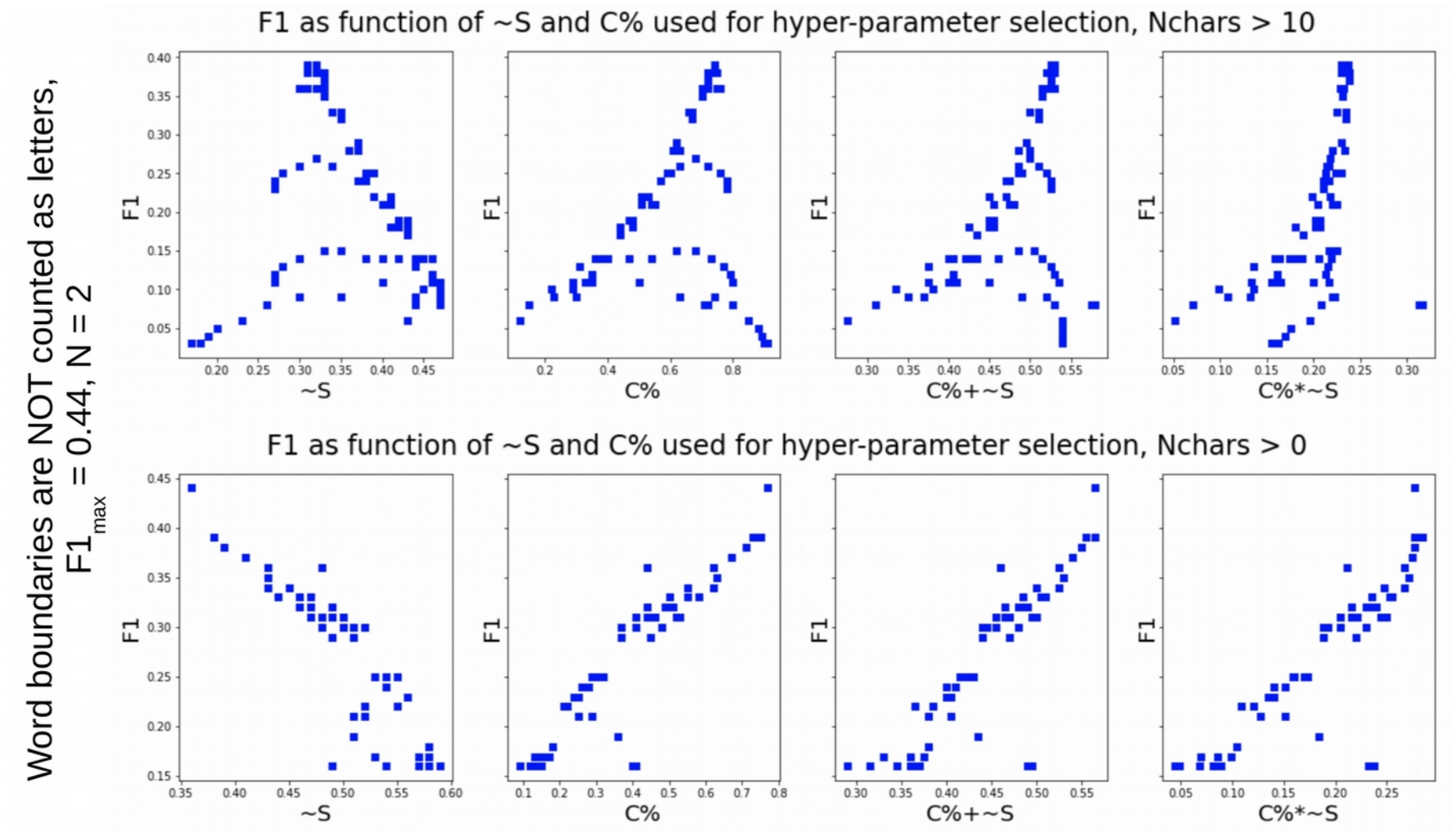

**Fig. 5.** Scatter plots representing distribution of F1 scores of sub-word segmentation compared to greedy morphological parsing relying on dictionary of English suffixes and prefixes against culture-agnostic metrics such as (left to right) anti-entropy, compression factor, average of the two, and production of the two. Top row corresponds to experiment based on English lexicon words longer than 10 characters, bottom row corresponds to use of all words in the lexicon.

### 3.2 Morphological parsing or subword-level text segmentation

Preliminary cursory study for morphological parsing or subword segmentation has been performed for English lexicon. The goal of the study was to see if accuracy of such segmentation validated on real English morphology can be associated with the culture-agnostic metric described earlier. We have have trained the graph model of N-gram-to-character transitions with *N* in range from 1 to 10 based on English lexicon mentioned in section 2.2 with account to relative word frequency known from the lexicon data. Then we have performed subword segmentation with different hyper-parameters such as *N* and threshold on transition freedom peak value, used to detect the text segment boundary, according to [9,11], using the same lexicon as a test set. The F1 score of segmentation was computed on word-by-word basis referring to “greedy” morphological parser relying on reference dictionaries of English prefixes and suffixes. The overall F1 score across words was computed as weighted average with account to word frequency in lexicon. Similarly, the anti-entropy $\tilde{S}$ and compression factor $C\%$ metrics were computed with account to the word frequency. In addition to the basic metrics we also used average $\tilde{S}+C\%$ and production $\tilde{S}*C\%$ derivatives.

Results of the experiment in a space of hyper-parameters are presented on Fig.1 are showing strong positive correlation between morpho-parsing F1 score and compression factor, as it was found for the case of tokenization earlier. However, surprisingly, the anti-entropy has been found to render rather strongly negative correlation with the F1 score, quite opposite to what we have learned in case of tokenization, as it is shown on Fig. 6.

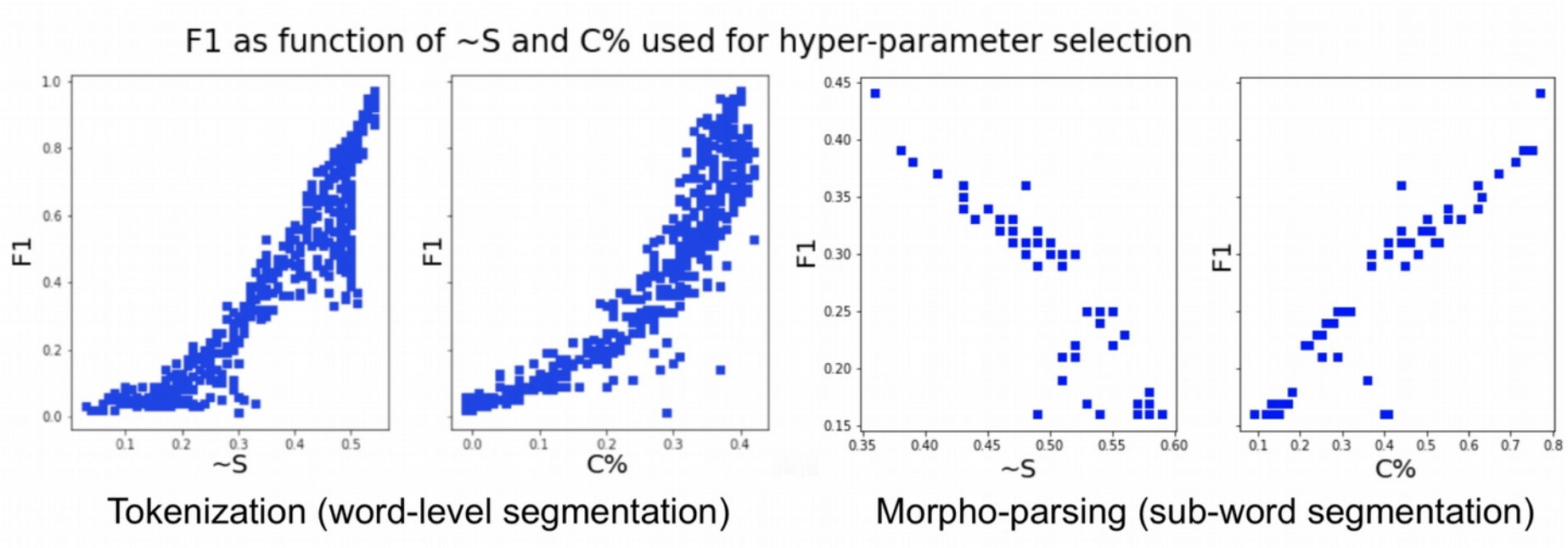

**Fig. 6.** Illustration of similarly positive connection between text segmentation F1 score and compression factor $C\%$ with opposite nature of connection of F1 score and anti-entropy $\tilde{S}$ in case of tokenization (left two scatter plots) and morpho-parsing (right two scatter plots).

## 4 Conclusions

We have confirmed strongly positive connection between the culture-agnostic information metrics such as anti-entropy, compression factor and cross-cultural consistency (cross-split F1 score) as guiding the language model structure at the level of word-level segmentation or tokenization across Russian, English and Chines languages. That suggests the nature of the language evolution as a generic process of development of symbolic communication codes efficient from multiple perspectives and respective measures.

We have also found similar connection in respect to sub-word level segmentation or morphological parsing in case of English, but that was limited to compression factor measure, while the association of the language structure at this level with anti-entropy has been found opposite, so the most accurate morphological segmentation of a text corresponds to maximized compression factor and maximized entropy (minimized anti-entropy) at the same time which appears surprising, needs further study.

## 5 Acknowledgments

We are grateful to Sergey Terekhov and Nikolay Mikhaylovskiy for valuable questions, critique, recommendations and suggestions during the course of work.